\documentclass[11pt]{article}

\usepackage[final]{acl}

\usepackage{times}
\usepackage{latexsym}
\usepackage[T1]{fontenc}
\usepackage[utf8]{inputenc}
\usepackage{microtype}
\usepackage{inconsolata}
\usepackage{graphicx}
\usepackage{booktabs}
\usepackage{amsmath}
\usepackage{amssymb}

\title{Below the Noise Floor: Bimodal Seed Collapse and \\ Distinct Failure Modes in Small-Model Knowledge Distillation}

\author{
  Dipto Sumit \quad Sakib Ul Haque \quad Farig Sadeque \\
  Department of Computer Science and Engineering \\
  BRAC University, Bangladesh \\
  \texttt{\{dipto.sumit, sakib.ul.haque\}@g.bracu.ac.bd} \\
  \texttt{farig.sadeque@bracu.ac.bd}
}

\begin{document}
\maketitle

\begin{abstract}
Function routing --- selecting the correct API call from a fixed catalog given a natural-language request --- is a deployment problem where small students are attractive but knowledge distillation gains are typically reported single-seed, at scales where seed variance is unknown. On a 740-instance healthcare API routing task with a 1.5B Qwen student and a 20B teacher, we compare eight KD variants against supervised cross-entropy, using three to six seeds for key configurations. We find: (i) per-seed standard deviation ranges from 2.8 to 48.7 percentage points, swallowing every claimed KD gain below five points; (ii) three of seven KD variants exhibit \textbf{bimodal collapse}, with at least one in three to five seeds falling below 55\% accuracy while the others train normally, and a fourth showing elevated variance; (iii) collapse has distinct modes --- wrong-function selection for \texttt{ce\_Knowledge Distillation} and \texttt{ce\_paraphrase}, and a previously undocumented \textit{output-truncation} mode for \texttt{reasoning\_kd}, where the model emits reasoning but terminates before producing a function name (0.9\% accuracy); (iv) only \texttt{progressive\_kd} and \texttt{rank\_kd} avoid collapse across observed seeds, with $\sigma \leq 3.9$ pp; (v) a naive cross-split $+3.78$ pp gain from input enrichment reverses to $-2.70$ pp under controlled within-split multi-seed re-testing. Single-seed evaluation is therefore unable to detect central failure modes in small-model KD.
\end{abstract}

\section{Introduction}
\label{sec:intro}

Deploying small language models for function routing is attractive in latency- and cost-constrained settings, but these models frequently degrade under paraphrase variation and other forms of linguistic shift. The standard remedy is knowledge distillation \citep{hinton2015distilling}: a growing literature proposes increasingly sophisticated objectives (confidence-gated KD, selective distillation, contrastive distillation, progressive curricula), each reporting improvements of 1--5 percentage points over a supervised baseline. \textbf{We argue these gains are, in large part, artifacts of single-seed evaluation, and that the field is systematically missing two more interesting phenomena: KD-induced bimodal collapse, and qualitatively distinct collapse modes across seemingly related KD losses.}

On a 740-instance healthcare API routing task, we distill a 20B-parameter teacher into a 1.5B Qwen student and compare eight KD variants against a supervised cross-entropy baseline (\texttt{ce\_only}). Five conditions are run with four to six seeds; the remaining KD variants are reported with three seeds. We make five findings.

(i)~Per-seed standard deviation at this scale is large ($\sigma{=}2.8$--$48.7$ percentage points (pp) across conditions), comparable to or larger than the gaps between methods that the literature treats as meaningful.

(ii)~Three of the seven multi-seed KD configurations exhibit \textbf{bimodal collapse}: under identical data, model, and hyperparameters, at least one in three to five random seeds drives the student to below 55\% accuracy while the other seeds train to within a point or two of the baseline. A fourth (\texttt{gated\_kd}) shows elevated variance ($\sigma=10.5$ pp) without crossing the 55\% threshold in our sample. This is a qualitatively different failure mode from gradual degradation, and it is invisible to single-seed evaluation.

(iii)~Collapse is not a single phenomenon. We document at least two qualitatively distinct modes. \texttt{ce\_Knowledge Distillation} and \texttt{ce\_paraphrase} collapses produce \textit{wrong-function selection}: the model still outputs valid function names but selects the wrong one at high confidence. \texttt{reasoning\_kd} on seed 7 instead produces \textit{output truncation}: the model emits a structurally valid reasoning chain but terminates on a learned-substitute EOS token (a single CJK character, U+622A, meaning ``cut/intercept'') before producing a function name, yielding 0.9\% accuracy because it never commits to a prediction. The two modes have different implications for what is going wrong during training.

(iv)~Only \texttt{progressive\_kd} (delayed-onset KD, $\sigma=2.8$ pp over four seeds) and \texttt{rank\_kd} (order-only KD, $\sigma=3.9$ pp over four seeds) avoid collapse across all observed seeds. Both interventions share a property: they remove the full-strength KL gradient from the start of training, either by delaying it or by replacing it with a weaker pairwise margin signal.

(v)~A cross-split comparison appears to show a strong $+3.78$ pp gain from adding function descriptions to the prompt. A controlled within-split re-test with three seeds reverses this finding ($\Delta=-2.70$ pp, $\sigma=6$--$11$ pp), exposing it as a subset-selection artifact and providing a worked example of why single-seed cross-split evaluations produce unreliable headlines.

\paragraph{Our major contributions in this paper are as follows.}
\begin{itemize}
\item Eight KD variants versus \texttt{ce\_only} in a single controlled study, with three to six seeds per condition.
\item Identification of \textbf{bimodal seed collapse} as a failure mode of auxiliary-loss KD at small student scale, observed in three independent loss configurations (\texttt{ce\_Knowledge Distillation} (cross-entropy plus KL-divergence knowledge distillation), \texttt{ce\_paraphrase} (cross-entropy plus paraphrase-consistency loss), \texttt{reasoning\_kd}), with a fourth (\texttt{gated\_kd}) showing elevated variance but no sub-55\% seed in our sample.
\item Identification of a previously undocumented \textit{output-truncation} collapse mode in \texttt{reasoning\_kd}, qualitatively distinct from the wrong-function mode in \texttt{ce\_Knowledge Distillation}.
\item A paired worked example: naive cross-split shows $+3.78$ pp from input enrichment; controlled within-split shows $-2.70$ pp. The reversal demonstrates concretely why multi-seed reporting is needed.
\end{itemize}

\section{Related Work}
\label{sec:related}

\paragraph{Knowledge distillation.}
The standard formulation \citep{hinton2015distilling} trains a student to match the temperature-softened softmax distribution of a larger teacher, combining a hard cross-entropy term with a soft KL-divergence term. Extensions to language models include DistilBERT \citep{sanh2019distilbert} and sequence-level distillation \citep{kim-rush-2016-sequence}. More recent work distills reasoning capabilities via chain-of-thought rationales \citep{magister2023teaching, hsieh2023distilling}. \citet{stanton2021does} showed that students trained with KL-divergence often produce predictions that diverge from the teacher despite low loss, suggesting that the standard objective measures the wrong thing. Our work extends this skeptical line: we show that in small-data structured classification, KD-augmented objectives exhibit bimodal seed behavior --- normal training in most seeds, qualitatively distinct collapses in others --- that single-seed reporting cannot detect.

A separate line studies how to mitigate harmful teacher signal. Confidence-based gating modulates KD weight by teacher entropy; selective distillation \citep{wang2021selective} uses only high-confidence or high-agreement teacher predictions. Progressive or curriculum-style KD schedules \citep{jafari2021annealing} delay or anneal the teacher signal. We test variants from each of these families and find that the curriculum approach (delayed onset) is alone in eliminating collapse without sacrificing accuracy.

\paragraph{Function calling and tool use.}
Recent work has emphasized scaling the function catalog and reasoning about multi-step invocation. Gorilla \citep{patil2023gorilla} and ToolLLM \citep{qin2024toolllm} target catalogs of thousands of APIs; xLAM \citep{liu2024apigen} releases function-calling fine-tunes; the Berkeley Function Calling Leaderboard \citep{yan2024berkeley} has become a standard evaluation. Most of this work focuses on catalog breadth. We study the complementary regime: a fixed, modest catalog where the bottleneck is per-query routing accuracy under linguistic variation. \citet{chen2024octopus} and \citet{erdogan2024tinyagent} demonstrate that sub-2B models can achieve strong function-calling accuracy. Our methodological contribution is orthogonal: we ask whether the reported \textit{differences} between fine-tuning strategies at this scale are reliable.

\paragraph{Seed variance and small-data evaluation.}
\citet{dodge2020fine} showed that BERT fine-tuning on small datasets exhibits seed-to-seed standard deviations of several points on GLUE benchmarks and that random restarts substantially change conclusions. \citet{mosbach2021stability} and \citet{reimers2017reporting} make analogous arguments for sentence-level classifiers and sequence labeling. To our knowledge, no published comparison of small-model knowledge distillation variants has reported the bimodal collapse phenomenon we document, the qualitatively distinct collapse modes we identify, or per-seed variance in the magnitude we measure ($\sigma$ up to 48.7 pp).

\section{Setup and Methods}
\label{sec:methods}

\subsection{Task definition}
We study \textbf{candidate function selection}: given a natural-language user instruction $x$ and a candidate set of $K$ function names $\{f_1, \ldots, f_K\}$, predict the single correct function $f^* \in \{f_1, \ldots, f_K\}$. We do not study argument generation or multi-step tool use. $K$ ranges from 2 to 4 (mean 3.02).

\subsection{Data}
A routing benchmark constructed by the authors from publicly documented API schemas in the healthcare, insurance, and fitness domains. Instructions were generated by the authors as synthetic natural-language paraphrases of canonical intents covered by each API, with no real user queries or patient data involved (see also the Ethics Statement). The benchmark is partitioned by \textbf{paraphrase family}: distinct phrasings of the same intent are placed in the same split, so the test set contains rephrasings never seen during training. Two split variants are used: V1 splits (515 train / 110 dev / 111 test), and V2 splits (362 train / 81 dev / 78 test), where V2 is a subset of V1 restricted to instructions whose candidate functions all have catalog descriptions available.

\subsection{Student and teacher}
\paragraph{Student.} A 1.5B Qwen2.5-Instruct \citep{qwen25} fine-tuned with QLoRA \citep{dettmers2023qlora} ($r=32$, $\alpha=64$, dropout 0.05) targeting all linear projections. The base model is frozen at 4-bit precision; only LoRA adapters and a small ranking head are trained.

\paragraph{Teacher.} A 20B-parameter open-weight Mixture-of-Experts language model with native function-calling pretraining, fine-tuned on our training set. Teacher annotations are generated once and stored offline. The teacher achieves 83.1\% agreement with gold on V1 training samples.

\paragraph{Training.} Effective batch size 16 ($4\times4$ gradient accumulation), AdamW ($\beta_1{=}0.9$, $\beta_2{=}0.999$, $\epsilon{=}10^{-8}$) with cosine schedule and 5\% linear warmup, peak LR $2\times10^{-4}$, weight decay $0.01$, gradient clipping at 1.0, 10 epochs with early stopping (patience 3) on dev loss. Mixed precision (bfloat16) with gradient checkpointing on the frozen base model. The ranking head is a 2-layer MLP (hidden 512, ReLU, dropout 0.1) over mean-pooled final hidden states. Each training run took 35--50 minutes on a single RTX A6000 or comparable 16 GB consumer card; the full multi-seed sweep across eight conditions consumed approximately 30 GPU-hours.

\paragraph{Seeding.} The reported seed controls (a) numpy and PyTorch RNGs, (b) the LoRA adapter initialization, (c) the ranking-head initialization, and (d) the per-epoch data shuffle order. The same seed reproduces identical results bit-for-bit on the same hardware. Tokenization, the candidate-set order, and the teacher annotations are seed-independent (computed once offline).

\subsection{Student input representations}
\label{sec:prompts}
We used two input formats: \textbf{names-only} (candidate function names listed) and \textbf{descriptions} (each candidate annotated with a one-line description and typed parameter list). The main multi-seed comparison uses names-only on V1. Section~\ref{sec:controlled} runs a controlled within-split test toggling only this format.

\subsection{Teacher annotation methods}
We used two annotation procedures, both producing per-candidate soft scores. \textbf{Method A: JSON self-report} (used in the main conditions). The teacher outputs a JSON object containing \texttt{candidate\_scores}, a \texttt{chosen\_function}, and a 3-step \texttt{reasoning} chain. Mean entropy 0.385 nats. \textbf{Method B: log-probability harvest with temperature} (used only in the \texttt{logit\_kd} reference). Mean entropy 0.692 nats. We report \texttt{logit\_kd} as a cross-system reference rather than as a controlled comparison.

\subsection{Loss formulations}
All ten loss configurations we evaluate share a common skeleton $\mathcal{L}_{\text{total}} = \mathcal{L}_{\text{LM}} + \mathcal{L}_{\text{select}}$, where $\mathcal{L}_{\text{LM}}$ is causal-LM cross-entropy on the target string and $\mathcal{L}_{\text{select}}$ varies across configurations. The student includes a 2-layer MLP candidate ranking head over pooled hidden states. We compare ten configurations; $\mathbf{p}_T$ is the teacher's candidate distribution, $\mathbf{z}_S$ the student's candidate logits.

\paragraph{ce\_only.} $\mathrm{CE}(\mathbf{z}_S, y^*)$. Primary baseline.

\paragraph{ce\_Knowledge Distillation.} $\mathrm{CE} + \lambda_{KD} T^2\,\mathrm{KL}(\mathbf{p}_T \Vert \mathrm{softmax}(\mathbf{z}_S/T))$; $\lambda_{KD}{=}1.0$, $T{=}2$.

\paragraph{ce\_paraphrase.} $\mathrm{CE} + \lambda_{\text{para}}(1 - \cos(\mathbf{h}_i, \mathbf{h}_j))$ on in-batch paraphrase pairs; $\lambda_{\text{para}}{=}0.5$.

\paragraph{full.} $\mathrm{CE} + \lambda_{KD}\,\mathrm{KL}_T + \lambda_{\text{para}}(1{-}\cos)$.

\paragraph{rank\_kd.} CE plus pairwise margin over teacher-preferred candidate pairs: $\frac{1}{|P|}\sum_{(a,b)\in P}\max(0,\,m - (z_{S,a}{-}z_{S,b}))$, $m{=}0.5$. Only teacher ranking order matters.

\paragraph{gated\_kd.} CE plus KL weighted by teacher confidence: $(1-\tilde{H}(\mathbf{p}_T))\cdot\lambda_{KD}T^2\,\mathrm{KL}_T$.

\paragraph{progressive\_kd.} CE-only for epochs 1--4; KD weight linearly ramped epochs 5--7; full KD epochs 8--10.

\paragraph{reasoning\_kd.} KL applied only on samples where the teacher's argmax matches gold; CE only on disagreements.

\paragraph{contrastive\_kd.} InfoNCE \citep{oord2018representation} with paraphrase as positive, in-batch samples as negatives ($\tau{=}0.07$), plus CE on the candidate head.

\paragraph{logit\_kd.} Uses Method B annotations and a different ranking head. Loss: $0.5\,\mathcal{L}_{\text{LM}} + 0.5(\alpha\,\mathrm{KL}_T + \beta\,\mathrm{CE})$, $\alpha{=}0.7$, $\beta{=}0.3$.

\subsection{Evaluation}
\label{sec:eval}
On the held-out test split, we report three metrics: \textbf{Accuracy} (argmax of $\mathbf{z}_S$ vs gold); \textbf{Macro F1}; and \textbf{Paraphrase consistency (PC)}, the fraction of paraphrase families on which the model predicts the identical function name across all family members. The test split contains 14 such families. PC is strict (all-same) and noisy at this sample size; Section~\ref{sec:v1} flags one case where PC reaches 1.0 vacuously.

\section{Results}
\label{sec:results}

\subsection{Multi-seed V1 results: bimodal collapse}
\label{sec:v1}

\begin{table}[t]
\centering
\small
\setlength{\tabcolsep}{6pt}
\begin{tabular}{lccc}
\toprule
Method & $n$ & Mean $\pm \sigma$ & Collapse \\
\midrule
\texttt{ce\_only} & 6 & $82.6 \pm 4.8$ & --- \\
\texttt{progressive\_kd} & 4 & $82.5 \pm 2.8$ & none \\
\texttt{rank\_kd} & 4 & $80.9 \pm 3.9$ & none \\
\texttt{full} & 4 & $78.2 \pm 6.6$ & none \\
\texttt{ce\_paraphrase} & 5 & $75.7 \pm 13.4$ & 1/5 \\
\texttt{gated\_kd} & 4 & $73.9 \pm 10.5$ & none \\
\texttt{ce\_Knowledge Distillation} & 5 & $61.1 \pm 22.1$ & 2/5 \\
\texttt{reasoning\_kd} & 3 & $57.1 \pm 48.7$ & 1/3$^\dagger$ \\
\bottomrule
\end{tabular}
\caption{Multi-seed V1 test accuracies (\%), names-only prompts, ranked by mean. Collapse = number of seeds with test accuracy $<55$\%. Three of seven KD configurations exhibit collapse; a fourth (\texttt{gated\_kd}) shows elevated variance without a sub-55\% seed. Only \texttt{progressive\_kd} and \texttt{rank\_kd} are stable across all observed seeds. Per-seed values in Appendix~\ref{app:per_seed}. $^\dagger$\texttt{reasoning\_kd} seed 7 collapses to 0.9\% via a qualitatively distinct \textit{output-truncation} mode (Section~\ref{sec:truncation}), not wrong-function selection.}
\label{tab:v1_multiseed}
\end{table}

Table~\ref{tab:v1_multiseed} reports the eight conditions for which we ran three or more seeds. Per-seed values are in Appendix~\ref{app:per_seed}.

\paragraph{Bimodal collapse in auxiliary-loss KD.} \texttt{ce\_Knowledge Distillation} collapses on 2 of 5 seeds (31.5\%, 48.7\%), with the remaining three seeds (62.2, 81.1, 82.0) clustered near or slightly below baseline. Its standard deviation, $\sigma=22.1$ pp, is more than four times the baseline's. \texttt{ce\_paraphrase} shows the same pattern at lower frequency: one seed at 52.3\%, four seeds between 77 and 86\%. \texttt{reasoning\_kd} produces the most extreme collapse, with one of three seeds at 0.9\% --- a qualitatively different failure mode we examine separately in Section~\ref{sec:truncation}. The other auxiliary-loss configurations (\texttt{gated\_kd}, \texttt{full}) show elevated variance ($\sigma=10.5$ and 6.6) without sub-55\% seeds in our sample, suggesting milder versions of the same instability. This is invisible to single-seed evaluation: depending on which seed is drawn, the same training recipe would be reported as ``ties with baseline'' or ``catastrophic failure.''

\paragraph{Two methods stable across all seeds.} \texttt{progressive\_kd} (delayed KD onset) achieves $\sigma=2.8$ pp across four seeds (79.3--85.6), with no seed below 79\% accuracy. \texttt{rank\_kd} (order-only KD) achieves $\sigma=3.9$ pp across four seeds (77.5--84.7). Both interventions share a structural property: they remove the full-strength KL-divergence gradient from the start of training, either by delaying it (\texttt{progressive\_kd}) or by replacing it with a weaker pairwise margin signal that uses only teacher rankings, not magnitudes (\texttt{rank\_kd}). Neither significantly beats the baseline on mean accuracy ($-0.1$ and $-1.7$ pp), but both achieve sub-baseline variance.

\paragraph{Method ranking depends on the seed.} Under seed 42, the top three are \texttt{ce\_only}/\texttt{full}/\texttt{progressive\_kd} clustered at 85.6\%. Under seed 123, \texttt{ce\_Knowledge Distillation} (82.0) and \texttt{ce\_paraphrase} (80.2) outrank \texttt{ce\_only} (75.7). Under seed 7, \texttt{reasoning\_kd} drops to 0.9\% while \texttt{rank\_kd} reaches 83.8\%. No two seeds produce the same ranking. Single-seed comparison of these methods is, by direct measurement, uninformative about which is better.

\subsection{Distinct collapse modes}
\label{sec:truncation}

\begin{figure*}[t]
\centering
\includegraphics[width=0.85\textwidth]{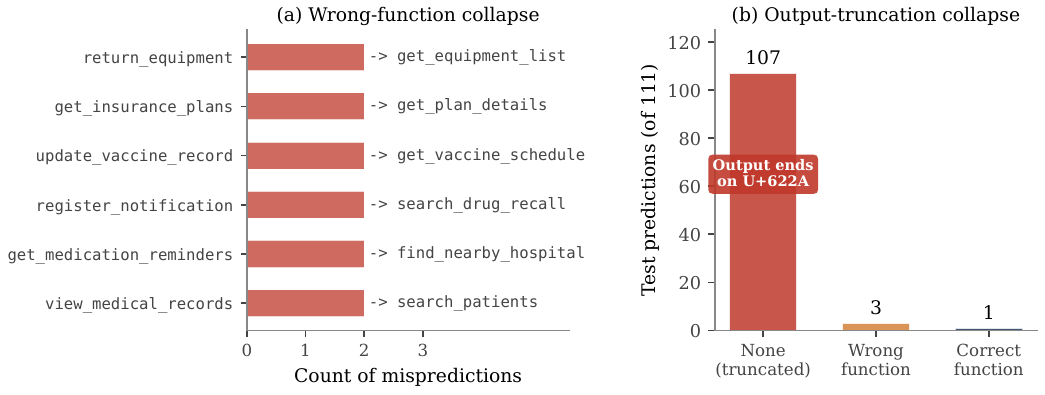}
\caption{Two qualitatively distinct collapse modes. (a) Wrong-function collapse in \texttt{ce\_Knowledge Distillation} seed 456 (31.5\% accuracy): the model emits valid function names but consistently aliases related intents (each row: gold $\to$ predicted, two test instances per row). (b) Output-truncation collapse in \texttt{reasoning\_kd} seed 7 (0.9\% accuracy): 107 of 111 test predictions return \texttt{None} because the output terminates on a learned single-character EOS surrogate (Unicode codepoint U+622A, meaning ``cut/intercept'') before the model commits to a function name.}
\label{fig:collapse_modes}
\end{figure*}

The collapses in Table~\ref{tab:v1_multiseed} are not a single phenomenon. We identify at least two qualitatively distinct modes by inspecting raw outputs from the collapsed checkpoints (Figure~\ref{fig:collapse_modes}).

\paragraph{Wrong-function selection (\texttt{ce\_Knowledge Distillation}, \texttt{ce\_paraphrase}).} On collapsed seeds, the trained student emits structurally valid outputs --- a reasoning chain followed by a function name --- but selects the wrong function with high confidence. Per-class confusion matrices for the worst \texttt{ce\_Knowledge Distillation} seed (31.5\%) show errors concentrated on a small subset of intent families: \texttt{return\_equipment} predicted as \texttt{get\_equipment\_list}, \texttt{view\_medical\_records} as \texttt{search\_patients}, \texttt{add\_remedy} as \texttt{search\_remedies}. The student has learned a degenerate mapping that aliases related-but-distinct intents.

\paragraph{Output truncation (\texttt{reasoning\_kd}).} On seed 7, \texttt{reasoning\_kd} collapses to a different mode entirely. The model produces 1 correct prediction out of 111. Inspection of the raw outputs reveals that 107 of the 110 wrong predictions are \texttt{None} --- the regular expression that extracts the function name from the reasoning chain returns no match. The model outputs a structurally valid two-step reasoning chain (\texttt{reasoning\_coverage}=1.0, \texttt{avg\_reasoning\_steps}=2.07) and then terminates on a single CJK character (Unicode codepoint U+622A, meaning ``cut'' or ``intercept'') before producing the required ``Chosen function:'' line. The character acts as a learned EOS substitute: the optimizer has converged on a fixed point that produces fluent-looking reasoning but never commits to a prediction. We note that \texttt{paraphrase\_consistency} for this collapsed seed is 1.0 --- vacuously, since every paraphrase variant hits the same truncation pattern.

\paragraph{Why this matters.} The two modes have different mechanistic implications. Wrong-function selection (panel a) is consistent with the model finding a degenerate minimum where teacher and gold signals partially cancel, producing high-confidence aliasing on intent-family clusters. Output truncation (panel b) is consistent with the optimizer learning to short-circuit the output sequence to avoid penalty entirely. Notably, \texttt{reasoning\_kd} --- which applies KL \textit{only} on teacher-gold-agreeing samples, and is therefore the variant with the least noisy KD signal in our study --- produces the most extreme failure. This refutes a simple ``teacher noise amplification'' explanation: it is not the disagreement-driven 16.9\% of training samples that destabilizes training, since removing them entirely makes things worse. We return to mechanism in Section~\ref{sec:why_multi_fails}, with the caveat that our data identify which interventions correlate with stability (delaying KD onset, replacing KL with a pairwise margin) but do not isolate a single causal mechanism.

\subsection{V2 results (descriptions prompt, single seed)}

\begin{table}[t]
\centering
\small
\begin{tabular}{lrrr}
\toprule
Method & Acc & F1 & PC \\
\midrule
\texttt{ce\_only} (baseline) & 88.5 & 79.2 & 90.9 \\
\texttt{ce\_Knowledge Distillation} & 85.9 & 72.9 & 72.7 \\
\texttt{ce\_paraphrase} & 84.6 & 69.4 & 81.8 \\
\texttt{full} & 69.2 & 52.5 & 45.5 \\
\bottomrule
\end{tabular}
\caption{V2 condition results, single seed ($n=78$ test). Given the seed variance documented for V1, single-point V2 numbers should be read as samples from high-variance distributions rather than stable per-method effects.}
\label{tab:v2}
\end{table}

Table~\ref{tab:v2} reports the four V2 single-seed runs. The 19 pp drop of \texttt{full} on V2 is consistent with a single-seed draw from the high-variance distribution characterized on V1 ($\sigma=6.6$, range 70.3--85.6); we do not interpret it as a stable per-method effect.

\subsection{The apparent input-enrichment effect}
\label{sec:naive}

\begin{table}[t]
\centering
\small
\begin{tabular}{lrrr}
\toprule
Condition & Acc & F1 & PC \\
\midrule
V1 \texttt{ce\_only} (names) & 84.7 & 74.0 & 78.6 \\
V2 \texttt{ce\_only} (descriptions) & 88.5 & 79.2 & 90.9 \\
\midrule
$\Delta$ (V2 $-$ V1), naive & \textbf{+3.78} & \textbf{+5.20} & \textbf{+12.34} \\
\bottomrule
\end{tabular}
\caption{Naive cross-split comparison. The apparent +3.78 pp gain confounds prompt format with subset selection (V2 is a description-available subset of V1).}
\label{tab:naive}
\end{table}

A naive cross-split comparison of \texttt{ce\_only} suggests that adding function descriptions to the student prompt produces a $+3.78$ pp gain, larger than any KD-design mean effect we measured. The comparison is confounded along two axes: prompt format and data subset. We initially read it as a real effect, then ran the controlled experiment below, and the apparent effect did not survive.

\subsection{Controlled within-split test}
\label{sec:controlled}

We fix the data split to V1 and vary only the prompt format. Three seeds (42, 123, 7) per arm, \texttt{ce\_only} loss, all other settings identical.

\begin{table}[t]
\centering
\small
\begin{tabular}{lcc}
\toprule
Arm & Acc (mean $\pm \sigma$) & $\Delta$ \\
\midrule
V1, names-only & $82.9 \pm 6.3$ & --- \\
V1, descriptions & $80.2 \pm 10.9$ & $-2.70$ \\
\bottomrule
\end{tabular}
\caption{Controlled within-split prompt-format comparison, 3 seeds per arm (per-seed values in Appendix~\ref{app:per_seed}). Descriptions show a $-2.70$ pp mean difference, with both arms exhibiting standard deviations larger than the difference. The naive $+3.78$ pp finding (Table~\ref{tab:naive}) does not survive.}
\label{tab:controlled}
\end{table}

\begin{figure*}[t]
\centering
\includegraphics[width=0.85\textwidth]{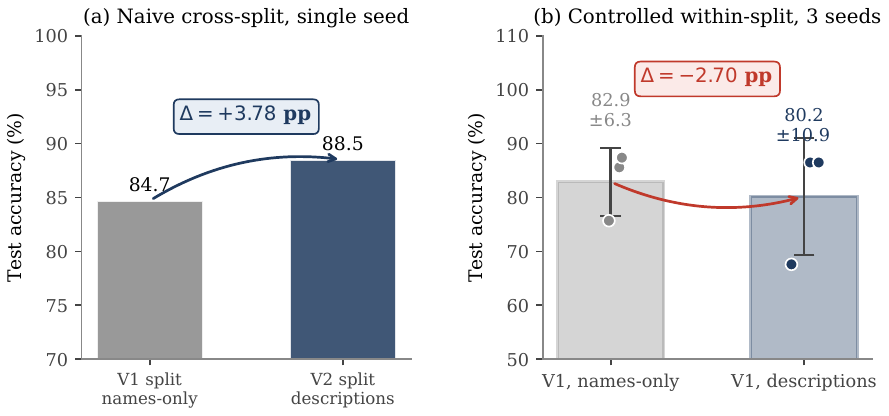}
\caption{The cross-split reversal. (a) A naive single-seed comparison across two splits suggests a $+3.78$ pp gain from adding function descriptions to the student prompt. (b) The same intervention re-tested under a controlled within-split protocol with three seeds per arm reverses sign ($-2.70$ pp). Dots are individual seed runs; bars and error bars show arm mean $\pm$ one standard deviation. Both arms' standard deviations are larger than the difference between their means: the apparent effect was a subset-selection artifact, and the within-split data do not support a description effect of either sign at this scale.}
\label{fig:reversal}
\end{figure*}

Two findings, methodologically central. First, the +3.78 pp apparent effect from the naive cross-split comparison \textit{reverses sign} under controlled within-split conditions ($-2.70$ pp; Figure~\ref{fig:reversal}), suggesting that the V2 subset is intrinsically easier --- not that the prompt format itself helps. Second, the per-seed variance is large: $\sigma=6.3$ for names-only and $\sigma=10.9$ for descriptions, with individual seed runs spanning 67.6\% to 87.4\% on the same 111-instance test split. The 2.70 pp arm-mean difference is well within both arms' standard deviations; we do not claim descriptions hurt either. We claim only that the data do not support a description effect of either sign at this scale.

\subsection{Summary of findings}

\begin{figure}[t]
\centering
\includegraphics[width=\columnwidth]{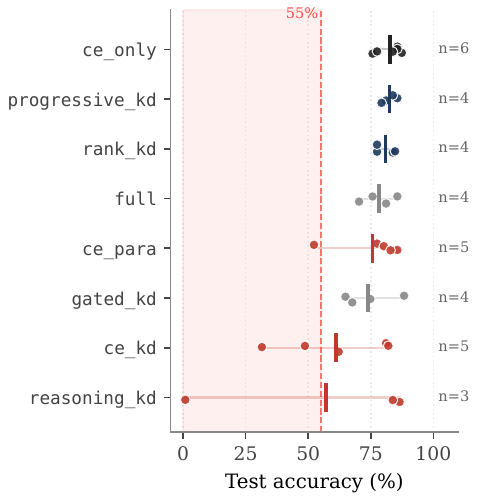}
\caption{Per-seed test accuracy (\%) for the multi-seed V1 conditions in Table~\ref{tab:v1_multiseed}. Dots are individual seed runs; vertical ticks show means. Sub-55\% seeds (the collapse region, dashed line) appear in three of seven KD configurations; \texttt{reasoning\_kd}'s 0.9\% point reflects the output-truncation mode (Section~\ref{sec:truncation}), not wrong-function selection. Only \texttt{progressive\_kd} and \texttt{rank\_kd} keep their full seed range above the baseline's worst seed.}
\label{fig:interventions}
\end{figure}

\begin{figure}[t]
\centering
\includegraphics[width=\columnwidth]{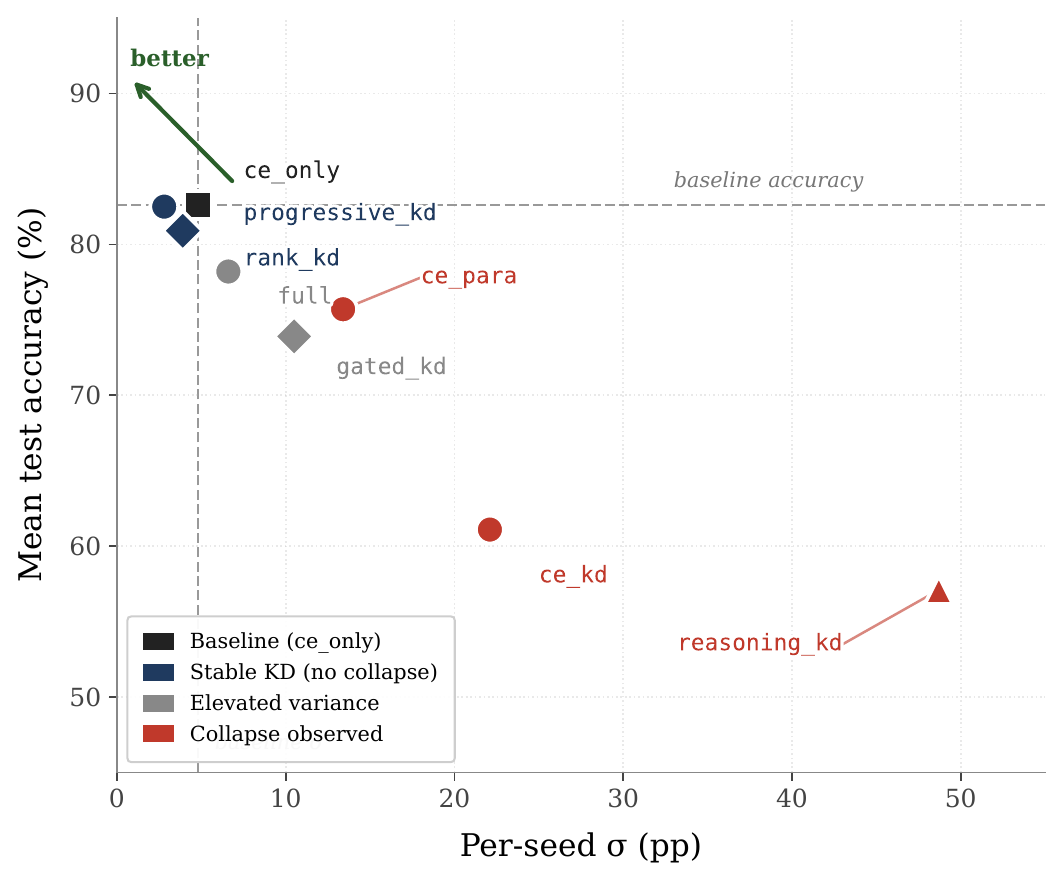}
\caption{Mean test accuracy versus per-seed standard deviation across the eight multi-seed conditions. Dotted reference lines mark the baseline (\texttt{ce\_only}). The stability-accuracy tradeoff is monotone in our data: methods that match the baseline mean (\texttt{progressive\_kd}, \texttt{rank\_kd}) sit at or below the baseline's variance; methods with elevated variance (\texttt{gated\_kd}, \texttt{ce\_paraphrase}) lose accuracy; methods with high variance lose substantially more (\texttt{ce\_Knowledge Distillation}, \texttt{reasoning\_kd}). No method achieves both higher mean and lower variance than the baseline.}
\label{fig:acc_vs_sigma}
\end{figure}

Five findings emerge: (i) per-seed variance dominates between-method differences at this scale (Figure~\ref{fig:acc_vs_sigma}); (ii) auxiliary-loss KD variants exhibit bimodal collapse on 1-in-3 to 2-in-5 seeds; (iii) collapse has at least two qualitatively distinct modes (wrong-function selection and output truncation; Figure~\ref{fig:collapse_modes}); (iv) only delayed-onset and order-only KD avoid collapse across all observed seeds; (v) a naive cross-split prompt-format gain reverses sign under controlled within-split re-testing (Figure~\ref{fig:reversal}).

\section{Discussion}
\label{sec:discussion}

\subsection{What the collapse modes suggest about mechanism}
\label{sec:why_multi_fails}

The most informative comparison in our data is between \texttt{ce\_Knowledge Distillation} and \texttt{reasoning\_kd}. Both apply full-strength KL-divergence between teacher and student distributions; they differ only in \textit{which} training examples receive the KD signal. \texttt{ce\_Knowledge Distillation} applies KL to every example, including the 16.9\% where the teacher disagrees with gold. \texttt{reasoning\_kd} applies KL only on the 83.1\% where teacher and gold agree --- so by construction, the KD signal it receives is denoised. If teacher noise were the primary driver of collapse, \texttt{reasoning\_kd} should be \textit{more} stable than \texttt{ce\_Knowledge Distillation}. Empirically, it is not: \texttt{reasoning\_kd} produces the most extreme collapse we observe (0.9\% accuracy, $\sigma$=48.7), and into a different failure mode than \texttt{ce\_Knowledge Distillation}. Teacher noise alone is not sufficient to explain the collapse pattern.

A second comparison provides positive evidence. \texttt{progressive\_kd} delays the KL gradient until epoch 5; \texttt{rank\_kd} replaces the KL gradient with a much weaker pairwise margin signal. Both avoid collapse entirely across all observed seeds. Both interventions share a structural property: they alter the timing or strength of the KL gradient at training onset. Combined with the \texttt{reasoning\_kd} null result, the natural hypothesis is that bimodal collapse is driven by early-training interaction between the full-strength KL gradient and the randomly-initialized ranking head, rather than by teacher noise per se.

\paragraph{Counterexamples that complicate the story.} Two observations resist this clean reading. First, \texttt{full} applies full-strength KL from epoch 1 alongside CE and paraphrase losses, yet did not collapse in our four seeds ($\sigma=6.6$, no sub-55\% seed). This may be a sample-size effect --- collapse rates of 1-in-5 to 1-in-3 in collapsing configurations leave a 30--50\% probability that four seeds would all train normally even if \texttt{full} shared the same failure distribution --- or it may indicate that mixing KL with other auxiliary objectives is genuinely protective. Second, \texttt{gated\_kd} applies confidence-weighted KL from epoch 1 and shows high variance ($\sigma=10.5$) without a sub-55\% seed in our sample. Both observations are consistent with collapse being a probabilistic outcome whose rate depends on a configuration-specific interaction we have not isolated. The two collapse modes (Figure~\ref{fig:collapse_modes}) further suggest the optimizer can be driven into more than one degenerate basin; isolating which basin a given seed-configuration combination will land in would require gradient-trajectory and loss-landscape measurements that we leave to future work.

\subsection{Practical implications}

For practitioners at this scale: treat \texttt{ce\_only} as a strong baseline; prefer \texttt{progressive\_kd} or \texttt{rank\_kd} when KD is desired (variance reduction without expected-mean cost); avoid auxiliary-loss configurations that apply full-strength KL from epoch 1, including agreement-conditioned variants; run at least three seeds before reporting improvements below 5 pp; inspect raw outputs on at least one seed, not just argmax accuracy, since at least one collapse mode in our data produces null predictions invisible to standard scoring.

\section{Conclusion}
\label{sec:conclusion}

We compared eight knowledge distillation variants against a supervised cross-entropy baseline on a 740-instance function routing task, with three to six seeds per condition. We find that auxiliary-loss KD variants exhibit \textbf{bimodal seed collapse} --- one in three to five random seeds drives the student below 55\% accuracy while the others train normally on identical data. Collapse is not a single phenomenon: \texttt{ce\_Knowledge Distillation} and \texttt{ce\_paraphrase} produce wrong-function selection, while \texttt{reasoning\_kd} produces a previously undocumented \textit{output-truncation} mode in which the trained model emits fluent reasoning but never commits to a function. Only \texttt{progressive\_kd} (delayed KD onset) and \texttt{rank\_kd} (order-only KD) avoid collapse across all observed seeds. The common property of the methods that succeed --- removing the full-strength KL gradient from the start of training --- and the failure of agreement-conditioning to prevent collapse together suggest that early-training KL interaction with the randomly-initialized ranking head, not teacher noise, drives the instability. An apparent $+3.78$ pp gain from a non-KD intervention reverses to $-2.70$ pp under controlled multi-seed re-testing. For the field, multi-seed reporting at small scale is not optional; bimodal collapse, including silent output-truncation modes, is a failure mode worth looking for in any small-model KD result.

\section*{Limitations}
\label{sec:limitations}

\paragraph{Uneven seed counts across conditions.} Seed counts are not equal across configurations (Table~\ref{tab:v1_multiseed}: $n=3$ to $n=6$). Experiments were run incrementally as compute became available over a four-day window, with seed allocation prioritized by the information each new run was expected to add: more seeds for the most-controlled comparisons (\texttt{ce\_only}, \texttt{progressive\_kd}), fewer for ablations that turned out to be qualitatively decisive on their first observation (\texttt{reasoning\_kd}, where seed 7's 0.9\% collapse arrived in the final batch). All seeds we ran are reported; none were dropped. The bimodal collapse and output-truncation findings are robust to the variation in $n$ --- they would survive any reasonable reweighting of the seeds we collected --- but precise $\sigma$ estimates for conditions with $n=3$, in particular \texttt{reasoning\_kd}'s $\sigma=48.7$, should be read as preliminary indicators of high variance rather than as tight population estimates. We recommend that follow-up work treat our $n=3$ values as load-bearing for the qualitative finding (collapse mode exists, distinct from wrong-function mode) but not for quantitative ranking against other conditions.

\paragraph{Three seeds is a low bar.} Even our best-sampled conditions ($n=6$) are below the seed counts \citet{dodge2020fine} and \citet{mosbach2021stability} recommend for stable BERT-fine-tuning variance estimates ($n \geq 10$). Our results therefore document the existence and qualitative shape of bimodal collapse but do not pin down the precise collapse-rate distribution for any single configuration.

\paragraph{Scope of claims.} A single healthcare/insurance/fitness domain with $\sim$700 training examples and a 1.5B student. Bimodal collapse may not transfer to larger catalogs, other domains, or datasets with $\sim10^5$+ examples. Our teacher has 16.9\% disagreement with gold; a more capable teacher could shift the comparison.

\paragraph{Mechanism is inferred from ablations, not measured directly.} The early-training KL-interaction hypothesis in Section~\ref{sec:why_multi_fails} is supported by two converging ablations (\texttt{progressive\_kd}, \texttt{rank\_kd}) and one informative null result (\texttt{reasoning\_kd}), but is complicated by two counterexamples (\texttt{full} and \texttt{gated\_kd}) and is not established by direct measurement of gradient trajectories or loss-landscape geometry.

\paragraph{PC metric.} Paraphrase consistency is computed over only 14 families and can reach 1.0 vacuously, as in the \texttt{reasoning\_kd} truncation case. Individual PC values should not be over-interpreted.

\paragraph{Output inspection is partial.} We manually inspected raw outputs only for the most extreme collapsed checkpoints. Other multi-seed conditions reporting wrong-function collapse may also contain subtler output pathologies we did not catch.

\section*{Ethics Statement}
\label{sec:ethics}

This work studies training stability of small language model students on a function routing task over public-style healthcare, insurance, and fitness API schemas. We do not collect new human data; the dataset consists of synthetic instruction-tool pairs without personal information. The healthcare-themed API schemas are illustrative templates, not connected to any real patient records or clinical systems, and our trained models are not intended for deployment in clinical or patient-facing settings without independent safety evaluation. All model and dataset artifacts were used in accordance with their stated licenses, terms of use, and redistribution restrictions

Our central finding --- that small-model knowledge distillation exhibits bimodal seed collapse including silent output-truncation failures --- has a direct safety implication: a single-seed deployment of any of the auxiliary-loss KD variants we test could, with non-trivial probability, produce a model that emits fluent reasoning but fails to commit to a function call, or that confidently routes user requests to the wrong API. In settings where the routed function has real-world consequences (medication look-up, appointment booking, claim submission), this is a deployment risk worth flagging. We recommend practitioners run multi-seed evaluation and inspect raw outputs --- not just argmax accuracy --- before shipping any KD-trained small student in a function-routing role.

Compute usage for the full multi-seed sweep is approximately 30 GPU-hours on a single consumer-grade 16 GB card. We release the training and evaluation code, per-seed result logs, and prompt templates with the paper to enable independent replication.

We used AI assistants (Claude) for writing assistance and coding support during the preparation of this work.

\bibliography{custom}

@misc{hinton2015distilling,
      title={Distilling the Knowledge in a Neural Network}, 
      author={Geoffrey Hinton and Oriol Vinyals and Jeff Dean},
      year={2015},
      eprint={1503.02531},
      archivePrefix={arXiv},
      primaryClass={stat.ML},
      url={https://arxiv.org/abs/1503.02531}, 
}

@article{sanh2019distilbert,
  title={DistilBERT, a distilled version of BERT: smaller, faster, cheaper and lighter},
  author={Victor Sanh and Lysandre Debut and Julien Chaumond and Thomas Wolf},
  journal={ArXiv},
  year={2019},
  volume={abs/1910.01108},
  url={https://api.semanticscholar.org/CorpusID:203626972}
}

@inproceedings{kim-rush-2016-sequence,
  title={Sequence-Level Knowledge Distillation},
  author={Kim, Yoon and Rush, Alexander M.},
  booktitle={Proceedings of the 2016 Conference on Empirical Methods in Natural Language Processing},
  pages={1317--1327},
  year={2016},
  address={Austin, Texas},
  publisher={Association for Computational Linguistics},
  doi={10.18653/v1/D16-1139},
  url={https://aclanthology.org/D16-1139/}
}

@inproceedings{magister2023teaching,
  title={Teaching Small Language Models to Reason},
  author={Magister, Lucie Charlotte and Mallinson, Jonathan and Adamek, Jakub and Malmi, Eric and Severyn, Aliaksei},
  booktitle={Proceedings of the 61st Annual Meeting of the Association for Computational Linguistics (Volume 2: Short Papers)},
  pages={1773--1781},
  year={2023},
  address={Toronto, Canada},
  publisher={Association for Computational Linguistics},
  url={https://aclanthology.org/2023.acl-short.151/}
}

@inproceedings{hsieh2023distilling,
  title={Distilling Step-by-Step! Outperforming Larger Language Models with Less Training Data and Smaller Model Sizes},
  author={Hsieh, Cheng-Yu and Li, Chun-Liang and Yeh, Chih-Kuan and Nakhost, Hootan and Fujii, Yasuhisa and Ratner, Alexander and Krishna, Ranjay and Chen, Chen-Yu and Pfister, Tomas},
  booktitle={Findings of the Association for Computational Linguistics: ACL 2023},
  pages={8003--8017},
  year={2023},
  address={Toronto, Canada},
  publisher={Association for Computational Linguistics},
  url={https://aclanthology.org/2023.findings-acl.507/}
}

@inproceedings{stanton2021does,
  title={Does Knowledge Distillation Really Work?},
  author={Stanton, Samuel and Izmailov, Pavel and Kirichenko, Polina and Alemi, Alexander A. and Wilson, Andrew Gordon},
  booktitle={Advances in Neural Information Processing Systems},
  volume={34},
  pages={6906--6919},
  year={2021},
  url={https://proceedings.neurips.cc/paper/2021/hash/376c6b9ff3bedbbea56751a84fffc10c-Abstract.html}
}

@inproceedings{wang2021selective,
  title={Selective Knowledge Distillation for Neural Machine Translation},
  author={Wang, Fusheng and Yan, Jianhao and Meng, Fandong and Zhou, Jie},
  booktitle={Proceedings of the 59th Annual Meeting of the Association for Computational Linguistics and the 11th International Joint Conference on Natural Language Processing (Volume 1: Long Papers)},
  pages={6456--6466},
  year={2021},
  address={Online},
  publisher={Association for Computational Linguistics},
  url={https://aclanthology.org/2021.acl-long.504/}
}

@inproceedings{jafari2021annealing,
    title = "Annealing Knowledge Distillation",
    author = "Jafari, Aref  and
      Rezagholizadeh, Mehdi  and
      Sharma, Pranav  and
      Ghodsi, Ali",
    editor = "Merlo, Paola  and
      Tiedemann, Jorg  and
      Tsarfaty, Reut",
    booktitle = "Proceedings of the 16th Conference of the European Chapter of the Association for Computational Linguistics: Main Volume",
    month = apr,
    year = "2021",
    address = "Online",
    publisher = "Association for Computational Linguistics",
    url = "https://aclanthology.org/2021.eacl-main.212/",
    doi = "10.18653/v1/2021.eacl-main.212",
    pages = "2493--2504",
}

@misc{patil2023gorilla,
      title={Gorilla: Large Language Model Connected with Massive APIs}, 
      author={Shishir G. Patil and Tianjun Zhang and Xin Wang and Joseph E. Gonzalez},
      year={2023},
      eprint={2305.15334},
      archivePrefix={arXiv},
      primaryClass={cs.CL},
      url={https://arxiv.org/abs/2305.15334}, 
}

@inproceedings{qin2024toolllm,
title={Tool{LLM}: Facilitating Large Language Models to Master 16000+ Real-world {API}s},
author={Yujia Qin and Shihao Liang and Yining Ye and Kunlun Zhu and Lan Yan and Yaxi Lu and Yankai Lin and Xin Cong and Xiangru Tang and Bill Qian and Sihan Zhao and Lauren Hong and Runchu Tian and Ruobing Xie and Jie Zhou and Mark Gerstein and dahai li and Zhiyuan Liu and Maosong Sun},
booktitle={The Twelfth International Conference on Learning Representations},
year={2024},
url={https://openreview.net/forum?id=dHng2O0Jjr}
}

@article{liu2024apigen,
  title={{APIGen}: Automated Pipeline for Generating Verifiable and Diverse Function-Calling Datasets},
  author={Liu, Zuxin and Hoang, Thai and Zhang, Jianguo and Zhu, Ming and Lan, Tian and Kokane, Shirley and Tan, Juntao and Yao, Weiran and Liu, Zhiwei and Feng, Yihao and others},
  journal={arXiv preprint arXiv:2406.18518},
  note={NeurIPS 2024 Datasets and Benchmarks},
  year={2024}
}

@article{yan2024berkeley,
  title={Berkeley Function Calling Leaderboard},
  author={Yan, Fanjia and Huang, Huanzhi and Wang, Sida and Mao, Qianli and Zhang, Tianjun and and others},
  journal={arXiv preprint arXiv:2407.01536},
  year={2024}
}

@article{chen2024octopus,
  title={Octopus v2: On-device language model for super agent},
  author={Chen, Wei and Li, Zhiyuan},
  journal={arXiv preprint arXiv:2404.01744},
  year={2024}
}

@inproceedings{erdogan2024tinyagent,
  title={{TinyAgent}: Function Calling at the Edge},
  author={Erdogan, Lutfi Eren and Lee, Nicholas and Jha, Siddharth and Kim, Sehoon and Tabrizi, Ryan and Moon, Suhong and Hooper, Coleman and Anumanchipalli, Gopala and Keutzer, Kurt and Gholami, Amir},
  booktitle={Proceedings of the 2024 Conference on Empirical Methods in Natural Language Processing: System Demonstrations},
  pages={80--88},
  year={2024},
  address={Miami, Florida, USA},
  publisher={Association for Computational Linguistics},
  url={https://aclanthology.org/2024.emnlp-demo.9/}
}

@misc{dodge2020fine,
      title={Fine-Tuning Pretrained Language Models: Weight Initializations, Data Orders, and Early Stopping}, 
      author={Jesse Dodge and Gabriel Ilharco and Roy Schwartz and Ali Farhadi and Hannaneh Hajishirzi and Noah Smith},
      year={2020},
      eprint={2002.06305},
      archivePrefix={arXiv},
      primaryClass={cs.CL},
      url={https://arxiv.org/abs/2002.06305}, 
}

@inproceedings{mosbach2021stability,
  title={On the Stability of Fine-tuning {BERT}: Misconceptions, Explanations, and Strong Baselines},
  author={Mosbach, Marius and Andriushchenko, Maksym and Klakow, Dietrich},
  booktitle={International Conference on Learning Representations},
  year={2021},
  url={https://openreview.net/forum?id=nzpLWnVAyah}
}

@inproceedings{reimers2017reporting,
  title={Reporting Score Distributions Makes a Difference: Performance Study of {LSTM}-networks for Sequence Tagging},
  author={Reimers, Nils and Gurevych, Iryna},
  booktitle={Proceedings of the 2017 Conference on Empirical Methods in Natural Language Processing},
  pages={338--348},
  year={2017},
  address={Copenhagen, Denmark},
  publisher={Association for Computational Linguistics},
  url={https://aclanthology.org/D17-1035/}
}

@article{dettmers2023qlora,
  title={{QLoRA}: Efficient Finetuning of Quantized {LLMs}},
  author={Dettmers, Tim and Pagnoni, Artidoro and Holtzman, Ari and Zettlemoyer, Luke},
  journal={arXiv preprint arXiv:2305.14314},
  note={NeurIPS 2023},
  year={2023}
}

@misc{oord2018representation,
      title={Representation Learning with Contrastive Predictive Coding}, 
      author={Aaron van den Oord and Yazhe Li and Oriol Vinyals},
      year={2019},
      eprint={1807.03748},
      archivePrefix={arXiv},
      primaryClass={cs.LG},
      url={https://arxiv.org/abs/1807.03748}, 
}

@misc{qwen25,
      title={Qwen2.5 Technical Report},
      author={An Yang and Baosong Yang and Beichen Zhang and Binyuan Hui and Bo Zheng and Bowen Yu and Chengyuan Li and Dayiheng Liu and Fei Huang and Haoran Wei and Huan Lin and Jian Yang and Jianhong Tu and Jianwei Zhang and Jianxin Yang and Jiaxi Yang and Jingren Zhou and Junyang Lin and Kai Dang and Keming Lu and Keqin Bao and Kexin Yang and Le Yu and Mei Li and Mingfeng Xue and Pei Zhang and Qin Zhu and Rui Men and Runji Lin and Tianhao Li and Tianyi Tang and Tingyu Xia and Xingzhang Ren and Xuancheng Ren and Yang Fan and Yang Su and Yichang Zhang and Yu Wan and Yuqiong Liu and Zeyu Cui and Zhenru Zhang and Zihan Qiu},
      year={2025},
      eprint={2412.15115},
      archivePrefix={arXiv},
      primaryClass={cs.CL},
      url={https://arxiv.org/abs/2412.15115}
}
\newpage
\appendix

\section{Per-seed accuracy values}
\label{app:per_seed}

\begin{table}[h]
\centering
\small
\setlength{\tabcolsep}{4pt}
\begin{tabular}{lp{5.2cm}}
\toprule
Method & Per-seed test accuracy (\%) \\
\midrule
\texttt{ce\_only} & 87.4, 85.6, 75.7, 85.6, 83.8, 77.5 \\
\texttt{progressive\_kd} & 85.6, 81.1, 83.8, 79.3 \\
\texttt{rank\_kd} & 83.8, 77.5, 77.5, 84.7 \\
\texttt{full} & 85.6, 75.7, 70.3, 81.1 \\
\texttt{ce\_paraphrase} & 77.5, 80.2, 52.3, 85.6, 82.9 \\
\texttt{gated\_kd} & 74.8, 64.9, 67.6, 88.3 \\
\texttt{ce\_Knowledge Distillation} & 81.1, 82.0, 48.7, 31.5, 62.2 \\
\texttt{reasoning\_kd} & 86.5, 83.8, 0.9 \\
\midrule
\multicolumn{2}{l}{\textit{Controlled prompt-format comparison (Section~\ref{sec:controlled})}} \\
V1, names-only & 85.6, 75.7, 87.4 \\
V1, descriptions & 86.5, 67.6, 86.5 \\
\bottomrule
\end{tabular}
\caption{Per-seed accuracies for all multi-seed conditions. Seed assignments use the base set 42, 123, 7; additional seeds (456, 789, 1729, 2024, 2025) were added incrementally.}
\label{tab:per_seed_appendix}
\end{table}

\end{document}